\documentclass[10pt,twocolumn,letterpaper]{article}
\usepackage{iccv}
\usepackage{times}
\usepackage{epsfig}
\usepackage{graphicx}
\usepackage{amsmath}
\usepackage{amssymb}

\usepackage{microtype}
\usepackage{booktabs} 
\usepackage{multirow}
\usepackage{multicol}
\usepackage{color, colortbl}
\usepackage{adjustbox}
\usepackage{xspace}

\def\Mat#1{{\boldsymbol{#1}}}

\usepackage[pagebackref=true,breaklinks=true,letterpaper=true,colorlinks,bookmarks=false]{hyperref}

\iccvfinalcopy 

\ificcvfinal\fi

\begin{document}

\title{Query Adaptive Few-Shot Object Detection with Heterogeneous Graph Convolutional Networks}

\author{Guangxing Han, Yicheng He, Shiyuan Huang, Jiawei Ma, Shih-Fu Chang\\
Columbia University\\
{\tt\small \{gh2561,yh3330,sh3813,jiawei.m,sc250\}@columbia.edu}
}

\maketitle
\ificcvfinal\thispagestyle{empty}\fi

\begin{abstract}
   Few-shot object detection (FSOD) aims to detect never-seen objects using few examples. This field sees recent improvement owing to the meta-learning techniques by learning how to match between the query image and few-shot class examples, such that the learned model can generalize to few-shot novel classes. However, currently, most of the meta-learning-based methods perform pairwise matching between query image regions (usually proposals) and novel classes separately, therefore failing to take into account multiple relationships among them. In this paper, we propose a novel FSOD model using heterogeneous graph convolutional networks. Through efficient message passing among all the proposal and class nodes with three different types of edges, we could obtain context-aware proposal features and query-adaptive, multiclass-enhanced prototype representations for each class, which could help promote the pairwise matching and improve final FSOD accuracy. Extensive experimental results show that our proposed model, denoted as QA-FewDet, outperforms the current state-of-the-art approaches on the PASCAL VOC and MSCOCO FSOD benchmarks under different shots and evaluation metrics.
\end{abstract}

\section{Introduction}
\label{introduction}

With abundant annotated training examples of objects, deep neural networks are tailored to extract commonalities and detect object instances accordingly. 
However, such methods tend to over-fit when there are only a few examples available. On the other hand, having seen many similar objects, humans can recognize a novel object when shown only a few examples of it. 
Inspired by humans' outstanding ability to generalize knowledge, few-shot object detection aims to detect novel object instances in an image given a few examples of the novel object (a.k.a novel class) and abundant examples of other objects (a.k.a base classes).

\begin{figure}[t]
\begin{center}
\includegraphics[scale=0.28]{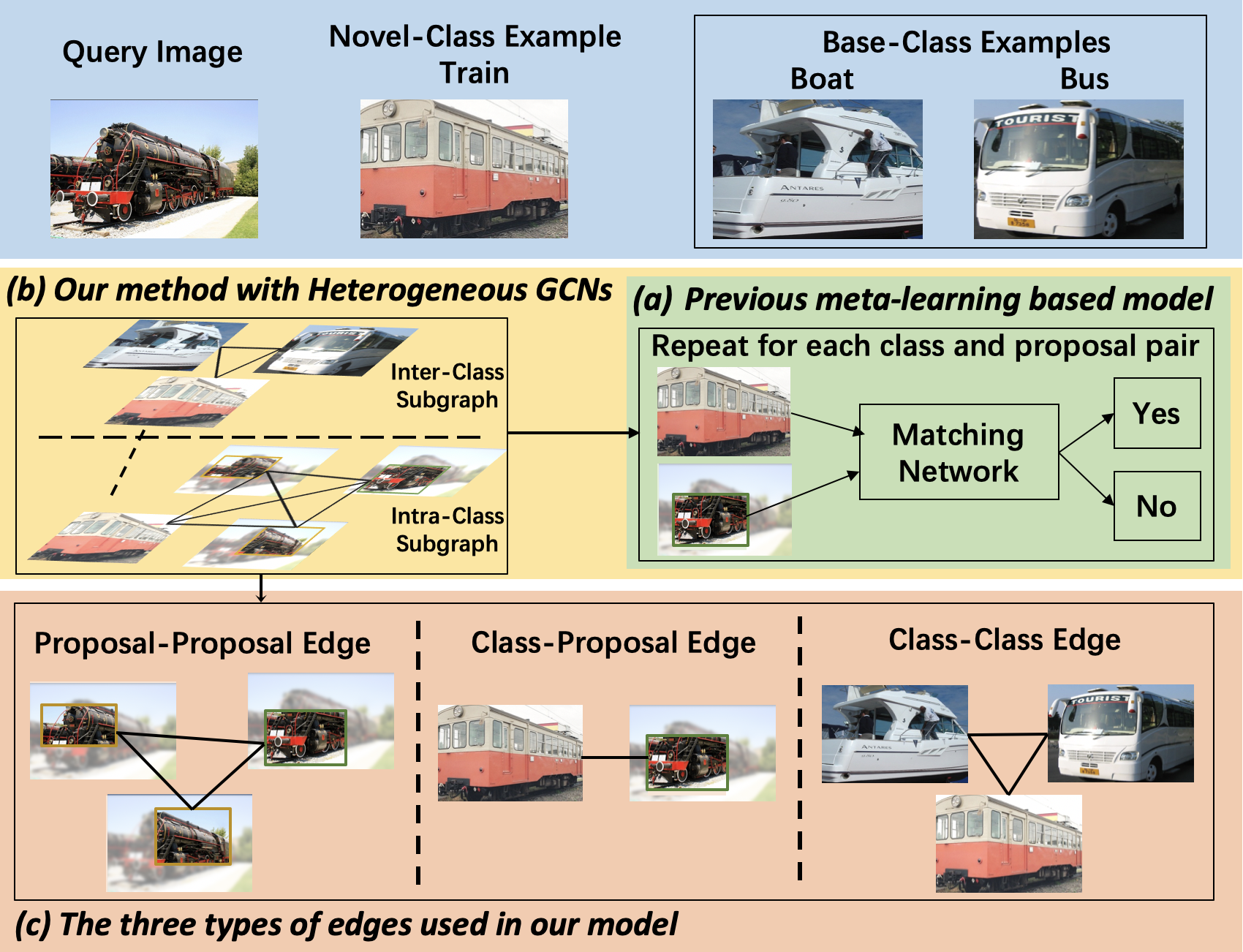}
\end{center}
\caption{Overview of our proposed few-shot object detection (FSOD) method. (a) Previous meta-learning-based FSOD method. These methods aim to learn how to match between the regions (usually proposals) in the query image and few-shot class examples. The matching network, the key module of these methods, will be repeatedly applied on each proposal and class pair. (b) Built upon the meta-learning-based approach, our method proposes a novel heterogeneous GCNs module with three different types of edges, allowing efficient message passing among all the nodes. (c) Three types of edges used in our model.}
\label{figure_1}
\end{figure}

Although object detection methods like Faster-RCNN \cite{ren2015faster} work extremely well on data-abundant base classes, it is non-trivial to adapt the model to few-shot novel classes. This is mainly because the softmax classifier used in R-CNN is tasked to perform classification among foreground classes and reject background regions at the same time. However, the notion of background changes when we adapt the model to novel classes. 
Typically, there are two approaches addressing this problem. One approach \cite{wang2020few,wu2020multi} is to use long-tailed learning methods for training on the unbalanced dataset, but the generalization ability of such model is still limited. The other approach \cite{fan2020few,perez2020incremental,wang2019meta,hsieh2019one} is to use meta-learning method to learn a class-agnostic few-shot detector on base classes, which can be easily adapted to novel classes without additional training. The simple framework and high detection accuracy make the meta-learning-based approach a promising choice for FSOD.

The key of these meta-learning approaches is to learn how to match between the regions (usually RPN proposals \cite{ren2015faster}) in the query image and few-shot class examples. This is achieved by learning a class-agnostic matching network with a binary classifier. The two inputs are the proposal feature extracted from an image \cite{Fast_R-CNN} and the prototype representation \cite{snell2017prototypical} of a few-shot class. The matching network will then be repeatedly applied on every proposal and class pair. However, there are three potential limits in this approach. \textbf{Firstly}, 
this kind of method turns out to be a ‘single-class’ detector, without modeling multi-class relations. This is important especially when there are base classes similar to the novel class as we could `borrow' robust features from these classes. 
\textbf{Secondly}, since the class prototype is extracted only from the few-shot examples, there could potentially be a huge discrepancy between the extracted prototype and the proposal feature, considering the wide variety of objects and their different image statistics.
\textbf{Thirdly}, proposals could be noisy and may not contain complete objects. Current methods do not consider the contextual information in the image for matching.

To address these three challenges, we propose a novel GCN-based FSOD model, denoted as QA-FewDet (\textbf{Q}uery-\textbf{A}daptive \textbf{Few}-Shot Object \textbf{Det}ection), which utilizes graph propagation to learn context-aware proposal features and query-adaptive, multiclass-enhanced class prototypes. 
As shown in Fig.\ \ref{figure_1}, 
we construct a graph among the proposal and class nodes for efficient class-class, class-proposal, and proposal-proposal communications. 
Firstly, by connecting different classes (including base classes) through \textbf{the class-class edge}, our method can model multi-class relations and enhance novel-class prototypes with prototypes from other similar classes. 
Secondly, \textbf{the class-proposal edge} provides mutual adaptation between class prototypes and proposal features and therefore reduces the distribution discrepancy between the two features. Meanwhile, it could provide additional examples of the class from proposals belonging to that class.
Thirdly, \textbf{the proposal-proposal edge} provides both local and global contextual information to help classification and bounding box localization.

The naive way of graph construction is to include all proposals and classes in a single graph. However, such a graph is memory-expensive and inefficient for message passing. To better incorporate these three types of edges in our model, as shown in Fig.\ \ref{figure_1}, 
we propose a novel heterogeneous graph consisting of a query-agnostic Inter-Class Subgraph, and multiple class-specific Intra-Class Subgraphs for each query image. The two subgraphs are processed sequentially for efficient message passing among all the nodes.

The entire network can be learned end-to-end using episode-based training on the abundant base-class data. 
To show the effectiveness of our model, we conduct comprehensive experiments on two widely-used FSOD benchmarks. Our model surpasses the current state-of-the-arts (SOTAs) by a huge margin under different shots and metrics.

Our contributions are: 1) To our knowledge, we are the first to propose a graph model that considers class-class, class-proposal, and proposal-proposal relations in few-shot object detection. 2) We propose a novel heterogeneous graph structure that allows efficient message passing among all the nodes. 3) Our model achieves significantly better results than current state-of-the-art methods on the PASCAL VOC and MSCOCO FSOD benchmarks across various settings.

\section{Related Works}

\textbf{Object Detection.}
Current DCNNs-based object detection methods can mainly be grouped into two categories: proposal-based methods and proposal-free methods. Proposal-based methods \cite{ren2015faster,he2017mask,R_RPN,han2018semi} divide object detection into two sequential stages by firstly generating a set of region proposals and then performing classification and bounding box regression for each proposal.
Proposal-free methods \cite{redmon2016you,liu2016ssd,tian2019fcos,SSD_TDR} directly predict the bounding boxes and the corresponding class labels on top of CNN features. We choose to use one of the most representative proposal-based methods, Faster R-CNN \cite{ren2015faster}, in our model as it usually has better detection accuracy than proposal-free detectors due to the cascade design and improving detection accuracy is still the top priority for FSOD.

\textbf{Few-Shot Learning and Meta-Learning.} Few-shot learning aims to recognize novel classes using only few examples. Meta-learning has been demonstrated a promising learning paradigm for few-shot learning tasks by transferring meta-knowledge learned from data-abundant base classes to data-scarce novel classes. Current meta-learning based few-shot learning methods can be roughly divided into three categories: optimization-based methods \cite{finn2017model}, parameter-generation-based methods \cite{gidaris2018dynamic} and metric-learning-based methods \cite{vinyals2016matching,snell2017prototypical,sung2018learning,ma2021PAL}.
Most of the few-shot learning methods are developed for the image classification task. 

\textbf{Few-Shot Object Detection}. Different from few-shot image classification, few-shot object detection needs to not only recognize objects with an arbitrary appearance, pose and scale using few-shot examples as references, but also localize (multiple) objects in the image and reject numerous background regions. Existing works can be mainly grouped into the following two categories: (1) Long-tailed-learning-based methods \cite{wang2020few,wu2020multi}. These methods attempt to learn object detection by using training data from both data-abundant base classes and data-scarce novel classes. To deal with the unbalanced training set, re-sampling \cite{wang2020few} and re-weighting \cite{lin2017focal} are the two main strategies \cite{kang2019decoupling}.
However, the models trained on the joint dataset are inflexible for adding never-seen few-shot classes. (2) Meta-learning-based methods \cite{karlinsky2019repmet,kang2019few}. Meta-learner \cite{kang2019few,fan2020few,perez2020incremental,yan2019meta} is introduced to acquire class-level meta knowledge via feature re-weighting and helps the model to generalize to novel classes. 
Meta-learning based methods \cite{wang2019meta,karlinsky2019repmet,xiao2020few,fan2020few,yang2020restoring,han2021meta,hsieh2019one,osokin2020os2d} has been demonstrated to be successful for FSOD.
Moreover, meta-learning based methods can be efficient for incrementally adding new few-shot classes during network inference. Our method belongs to this category.

\textbf{Graph Convolutional Networks}.
First proposed by \textit{Kipf et al.}\ \cite{kipf2016semi}, 
graph convolutional networks (GCNs) and their variants like Graph Attention Networks (GAT) \cite{velivckovic2017graph}, have seen massive applications in Computer Vision, including modeling video proposal relations in action localization \cite{nawhal2021activity, zeng2019graph}, object relations in visual relation reasoning \cite{mi2020hierarchical}, joint relations in skeleton-based action recognition \cite{yan2018spatial}, and object proposal relations in object detection \cite{chen2020hierarchical}. 
\textit{Liu et al.}\ \cite{liu2018structure} inject a GNN into a Faster R-CNN framework to contextualize the features of region proposals before the R-CNN classifier. This improves the results but it only shows results under traditional many-shot setting. Some methods \cite{wang2018zero} utilize GNNs on graphs that represent the ontology of concepts, which could enable generalization to unseen concepts by considering their relations with frequently-seen concepts. Different from previous works, we propose a novel heterogeneous GCNs in this paper that considers various relations within and between proposals and classes for FSOD.

\section{Task Formulation}

In few-shot object detection (FSOD), we split object classes $C$ into $C_{base}$ and $C_{novel}$ such that $C = C_{base} \cup C_{novel}$ and $C_{base} \cap C_{novel} = \emptyset$. For each class $c \in C$, its annotations $T_c$ contain the object instances' labels and bounding boxes inside the image. Formally, $T_c = \{(c, u, I) | u \in U, I \in \mathbb{R}^{H_I*W_I*3}\}$, and $U \subseteq \mathbb{R}^4 = \{(x, y, w, h)\}$ represents bounding boxes in the image.

For each class $c \in C_{base}$, we have plenty of annotated instances.
For each class $c \in C_{novel}$, we only have limited $K$ examples, also known as support images, (e.g., $|T_c| = K$ for $K = 1,5,10$). 
FSOD aims to detect novel-class object instances with few annotated object instances. 
Formally, given a query image $I_q \in \mathbb{R}^{H_q*W_q*3}$, FSOD outputs a set of detections $S_q = \{(c, u) | c \in C_{novel}, u \in U\}$.

\section{The Baseline FSOD Model} \label{backbone}

\begin{figure}[t]
\begin{center}
\includegraphics[scale=0.24]{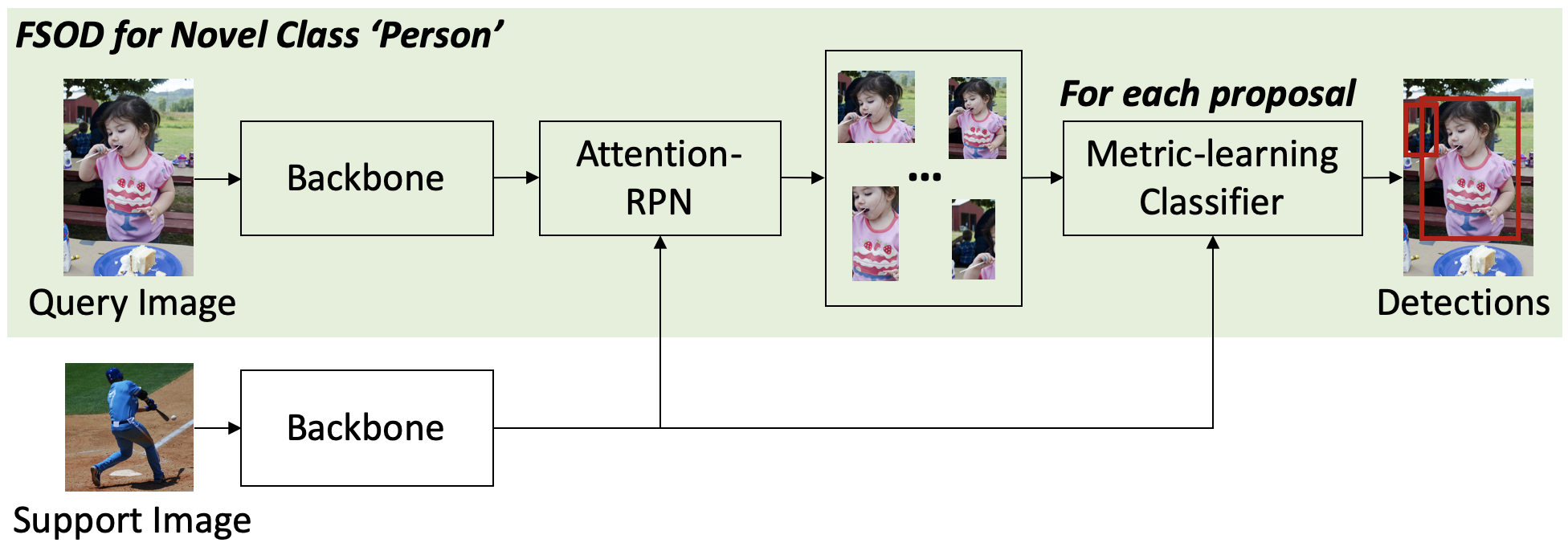}
\end{center}
\caption{The baseline FSOD model \cite{fan2020few}.}
\label{baseline}
\vspace{-5mm}
\end{figure}

As shown in Fig.\ \ref{baseline}, we mainly follow \textit{Fan et al.}'s work \cite{fan2020few} to build our baseline FSOD model, which adopts a siamese Faster R-CNN (with ResNet \cite{he2016deep}) with two branches. In one branch, a query image $I_q$ is fed into the feature extraction network to extract its feature $r(I_q)$ (output of $res4$ block) for the detection head. Similarly, the other branch extracts the feature $r(I_s)$ given an input support image $I_s$.

Then the Attention-RPN \cite{fan2020few} is used to produce $N$ ($N=100$ by default following \cite{fan2020few}) class-specific proposals $P_c = \{ \{ p_i^c\}_{i=1}^{N}, p_i^c \in U \}$ for the novel class $c \in C_{novel}$. After that, we use the $res5$ block and RoIPooling \cite{Fast_R-CNN} to extract the feature $f(p_i^c) \in \mathbb{R}^{H\times W\times C} (H=W=7, C=2048)$ for proposal $p_i^c$ from the query image feature $r(I_q)$.
We apply the same layers to $r(I_s)$, and take the average feature of all support images belonging to the novel class $c$ as the class prototype $f(c)$. After that, a multi-relation network \cite{fan2020few} is used to calculate similarity score between the proposal feature and the class prototype, and then produce the final detection results for class $c$ following \cite{ren2015faster}. The above process will be applied to each novel class independently.

The reason we use \cite{fan2020few} as our baseline model is as follows. Firstly, the whole framework is simple and elegant, and is a natural extension of the original Faster R-CNN to the few-shot setting. Secondly, as shown in the Section \ref{compare_with_SOTA}, it has achieved SOTA accuracy in major FSOD benchmarks.

Although \cite{fan2020few} has been demonstrated to be a promising FSOD model, there are three potential limits as discussed in the Section \ref{introduction}, and the primary cause is the separate classification of each proposal and class pair. To deal with this problem, we propose a novel FSOD model with heterogeneous GCNs in the following Section.

\section{Our Method with Heterogeneous GCNs}

The ultimate goal of our proposed heterogeneous GCNs is to enable efficient message passing among all the proposals and classes before pairwise classification. To this end, we first generate class-specific proposals and extract proposal features and class prototypes following the baseline model. Then we establish a novel heterogeneous graph using the generated proposal and class nodes, and use GCN layers to update features for each node in our graph in a sequential manner. After that, we use the updated features for the final pairwise classification.

\subsection{Overview of Our Heterogeneous Graph}

\begin{figure}[t]
\begin{center}
\includegraphics[scale=0.30]{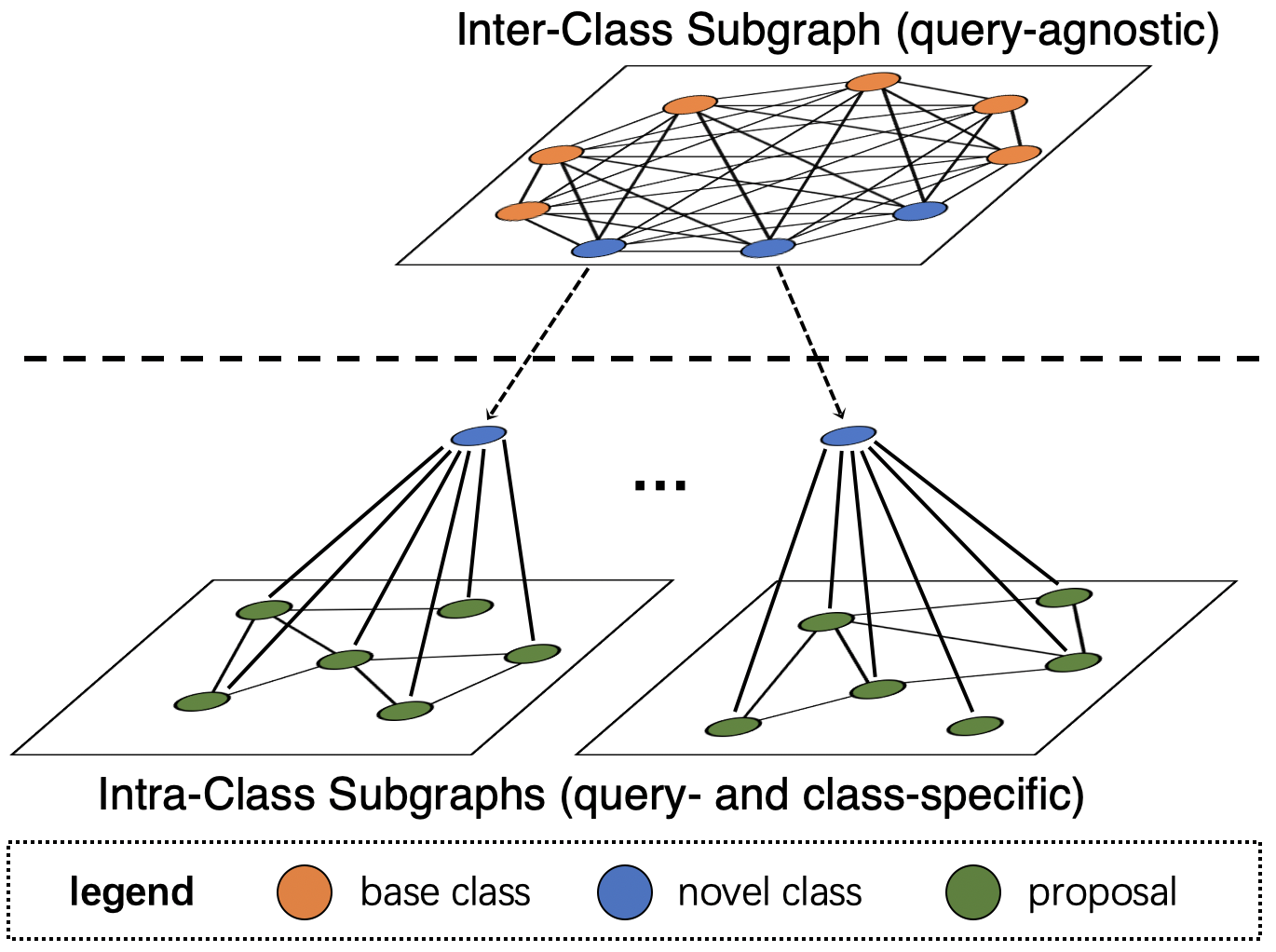}
\end{center}
\caption{Our proposed heterogeneous graphs.}
\label{figure_3}
\end{figure}

We aim to construct a graph to capture various types of relations within and between proposals and classes.
One way to construct the graph is to include all proposals and classes in a single graph. However, such a construction is memory-expensive as the number of proposals increases significantly with the number of classes, and the edge between cross-class proposals incurs redundant and noisy information.

Considering the class-specific proposals generated in our model, we build a heterogeneous graph $G$ that consists of two types of nodes, three types of edges, and two types of subgraphs as illustrated in Fig.\ \ref{figure_3}. 
Specifically, we build a graph $G=(V, E)$, where $V$ and $E$ denote the node sets and edge sets respectively. The two types of nodes in $V$ are the proposal nodes $V_p = \cup_{c \in C_{novel}} \{ P_c\}$, namely the class-specific proposals for each novel class, and the class nodes $V_c = C$ including all novel and base classes. $E$ has three components: the class-class edges $E_{c-c}$, the proposal-proposal edges $E_{p-p}$, and the class-proposal edges $E_{c-p}$. Our heterogeneous graph $G$ has two types of subgraphs: a query-agnostic Inter-Class Subgraph $G_{inter}$ shared among all query images, and multiple Intra-Class Subgraphs $G_{intra}=\{G_c, c\in C_{novel}\}$ for each query image, where $G_c$ represents the class-specific subgraph for novel class $c$.

\subsection{The Inter-Class Subgraph}

Inspired by previous works \cite{hinton2015distilling,gidaris2018dynamic}, modeling multi-class relationships could help enhance the prototype representation of few-shot classes and perform robust classification. 
In this paper, we first establish an Inter-Class Subgraph among all classes.
Given the set of all classes $C = \{c_1, c_2..., c_{|C|}\}$, we construct $G_{inter}=(V_{inter}, E_{inter})$, where $V_{inter} = C$ and $E_{inter} = \{(c_i, c_j) : 1 \leq i, j \leq |C|\}$. 
$A_{inter}$ is the adjacency matrix of $G_{inter}$ with edge weights for $E_{inter}$.
$G_{inter}$ is a graph with only class nodes, and considers relations among all base classes and novel classes for enhancing novel class prototypes. 

\textbf{The Class-Class Edge.} 
To efficiently aggregate valuable information from other classes, we compute the class-wise correlation between every class pair. The key idea is that if two classes are similar, it would be meaningful to update one class's prototype using that of the other class. Therefore, the edge weight between the two should increase correspondingly.
Formally, given a class pair $(c_i, c_j)$, we use cosine similarity to estimate their correlation by \footnote{As in Section \ref{backbone}, we use $f(c)$ to represent the prototype of class $c$. The same rule applies for the proposal feature.}

\begin{equation} \label{cosine_similarity}
e(c_i, c_j) = \frac{f(c_i)^T f(c_j)}{{\left\| f(c_i) \right\|_2 } \cdot {\left\| f(c_j) \right\|_2 }}
\end{equation}

We then apply softmax for each class $c_i$ to normalize its pairwise correlation
\begin{equation} \label{softmax}
A_{inter}^{ij} = \frac{exp(e(c_i, c_j))}{\sum_{k\in C} exp(e(c_i,c_k))},
\end{equation}

\subsection{The Intra-Class Subgraph}

Considering the class-specific proposals generated in our model, we build an Intra-Class Subgraph for each novel class. Each subgraph consists of one class node and the corresponding class-specific proposals.
Formally, for each novel class $c$, we construct $G_c = (V_c, E_c)$, where $V_c = \hat{P_c} \cup \{c \}$, and $ \hat{P_c} = P_c \cup \{g\}$. Here, $g$ denotes the `proposal' containing the entire image $I$, and we obtain its feature$f(g)$ by performing RoIPooling from the whole image feature. 
$A_c$ is the adjacency matrix of $G_c$ with edge weights for $E_c$.

To model the different types of relations within an Intra-Class Subgraph, we break $E_c$ into two components:

\textbf{The Proposal-Proposal Edge.} 
Contextual information has been demonstrated to be crucial for traditional many-shot object detection \cite{zeng2019graph,hu2018relation,liu2018structure}. In this paper, we apply this idea to the few-shot setting to obtain context-aware proposal features.
To be specific, we establish an edge between proposal $p_{i}$ \footnote{We use novel class $c$ as an example in the rest of this Section, and remove the class label $c$ in $p_{i}^c$ for simplicity reason.} and proposal $p_{j}$ if $\mathit{IoU}(p_{i}, p_{j}) > \theta$, where $\theta$ is a fixed threshold for determining meaningful overlaps ($\theta=0.7$ by default following \cite{zeng2019graph}), and $\mathit{IoU}$ is defined as

\begin{equation} \label{IoU}
\mathit{IoU}(p_{i}, p_{j}) = \frac{\cap(p_{i}, p_{j})}{\cup(p_{i}, p_{j})}, \quad p_{i}, p_{j} \in P_c
\end{equation}

At the same time, we also provide image-level contextual information by connecting the aforementioned global image node $g$ to each proposal.

By building edges with nearby proposals and with the global scene-context, the proposal-proposal edges enrich the original proposal features with both local and global contextual information, leading to better classification accuracy and more precise bounding box locations. 
The weight of the proposal-proposal edge in $A_c$ is calculated and normalized as in Eq. \ref{cosine_similarity} and Eq. \ref{softmax}. 

\textbf{The Class-Proposal Edge.} 
Traditional methods \cite{fan2020few,xiao2020few,yan2019meta} usually use the $K$-shot support images to extract the prototype of a novel class. However, the extracted prototype cannot represent the novel class well using very few examples. Moreover, it is challenging to extract robust novel-class prototypes using the feature backbone trained only on base classes.
To calibrate the different statistical distribution between the proposal feature and the class prototype, we introduce the class-proposal edge for dynamic mutual adaptations. Moreover, if the query image contains instances of the novel class, the class-specific proposals should cover these regions, and thus contributing extra-shots from the query image. This could help extract accurate novel-class prototype that is more suitable for the query image. 

In practice, we establish bidirectional edges between the class node $c$ and all the class-specific proposal nodes $P_c$. We show in Section \ref{ablation} that mutual adaptation is better than adaption with only a single direction.
Furthermore, since we only want to connect relevant proposals to the class node, we compute the cosine similarity as in Eq. \ref{cosine_similarity} to estimate the correlation between the class-proposal pair and filter out noisy relations. Then, for the class node $c$, we normalize the weights of all the incoming class-proposal edges similar to Eq. \ref{softmax}. For the proposal nodes, we use the original cosine similarity as the weight of the incoming class-proposal edge. 

\subsection{Our Heterogeneous GCNs}

After building our heterogeneous graph, we first perform message passing on the query-agnostic Inter-Class Subgraph for enhancing novel-class prototypes before processing any query image. With the enhanced novel-class prototypes, we then build multiple Intra-Class Subgraphs for each query image, and apply message passing for facilitating the communication among the proposal and class nodes.

For each graph, we sequentially perform $L$-layer GCNs ($L=1$ as shown in Section \ref{ablation}), which take in feature $X_0$ and output feature $X_L$ of the same size ($\mathbb{R}^{7\times 7\times 2048}$). In practice, we implement the $l^{th}$ GCN layer $(1 \leq l \leq L)$ by

\begin{equation} \label{GCN}
X^l = AX^{l-1}W^l
\end{equation}

In this equation, $X^l \in \mathbb{R}^{H*W*d_l}$ is the output feature of the $l^{th}$ GCN layer. $A_{inter}$ and $A_{c}$ are the respective adjacency matrix of the Inter-Class and class $c$'s Intra-Class Subgraph. $W^l \in \mathbb{R}^{d_{l-1}*d_l}$ is a learnable parameter matrix. Each layer is followed by a residual block.

In order to keep a consistent feature space between the class and proposal nodes, we apply the same number of learnable transformation layers to both of them as a siamese network.
Therefore, we do not use $W^l$ in any GCN layer in the Inter-Class Subgraph. A detailed discussion about this can be found in Section \ref{ablation}. 
For each class $c_i$, the effect of each layer is equivalent to a weighted sum of other class prototypes by \footnote{For simplicity reason, we use $f(c_i)$ and $\widetilde{f}(c_i)$ to denote the input and output of the $l^{th}$ layer GCN respectively. The similar strategy applies for Eq. \ref{Intra-Class Proposal} and Eq. \ref{Intra-Class Prototype}.}

\begin{equation} \label{Inter-Class GCN}
\widetilde{f}(c_i) = \sum_{j \in C} A_{inter}^{ji} \cdot f(c_j) + f(c_i)
\end{equation}

In the Intra-Class Subgraph, 
for each proposal $p_i$ of novel class $c$, let $P_{p_i} = \{g\} \cup \{p_j \in P_c \mid \mathit{IoU}(p_i, p_j) > \theta\} $ denote the set of its overlapping proposals and the global `proposal'. 
We aggregate $p_i$'s feature from the enhanced class prototype and proposal features of $P_{p_i}$ using a GCN layer by

\begin{equation} \label{Intra-Class Proposal}
\widetilde{f}({p_i}) = \big{(} A_{c}^{c{p_i}} \cdot f(c) + \sum_{p \in P_{p_i}} A_{c}^{p{p_i}} \cdot f(p)\big{)} W + f(p_{i}),
\end{equation}

$A_{c}^{c{p_i}}$ is the edge weight of class $c$ and proposal $p_i$ in $G_c$. $A_{c}^{p{p_i}}$ is defined similarly. We update class $c$'s prototype by

\begin{equation} \label{Intra-Class Prototype}
\widetilde{f}(c) = \big{(} f(c) + \sum_{p_{k} \in P_c} A_{c}^{{p_k}c} \cdot f(p_{i})\big{)}W + f(c),
\end{equation}

After obtaining the updated proposal features and the class prototype, we feed them into the pairwise matching network \cite{fan2020few} for the final classification.

\subsection{Training Framework}
To transfer knowledge from the base classes to the novel classes, we adopt a two-step training strategy.

\textbf{Meta-learning with Base Classes.}
With a pretrained feature extractor, we perform episode-based training on base classes. To simulate the few-shot scenario, each episode consists of one annotated query image and $K$ randomly sampled shots for each base class.
The whole model is supervised under a binary cross-entropy loss for classification and a smooth $L1$ loss for bounding box regression. During meta-testing, we can adapt our model to novel classes by simply calculating their prototype representations. 

\textbf{Fine-tuning with Novel Classes (Optional).}
We can further fine-tune the class-agnostic few-shot detector on novel classes following previous works \cite{fan2020few,xiao2020few,yan2019meta,wang2020few}. The difference between fine-tuning and meta-learning-only is that during fine-tuning, 
we use positive and negative proposals generated from the original novel class images to train our few-shot detector, 
while in meta-learning-only there is no training over novel classes. We study the performance of our model both with and without fine-tuning in Section \ref{ablation}.

\section{Experimental Results}

\subsection{Datasets}

We use two widely-used few-shot object detection benchmarks MSCOCO 2014 \cite{lin2014microsoft} and PASCAL VOC 2007 and 2012 \cite{everingham2010pascal} for model evaluation, and follow the same FSOD settings as previous works \cite{kang2019few,wang2020few} and use the same few-shot images for fair comparison.

\textbf{MSCOCO.} We set the 20 PASCAL VOC categories as novel classes and the remaining 60 categories as base classes. We use the same few-shot support images as \cite{kang2019few}. We report detection accuracy with AP, AP50, and AP75 under shots 1, 2, 3, 5, 10 and 30. 30-shot is considered as few-shot in MSCOCO dataset because the accuracy still largely falls behind the many-shot setting \cite{kang2019few}. We use the MSCOCO dataset for ablation study in Section \ref{ablation}.

\textbf{PASCAL VOC.} The 20 PASCAL VOC categories are split into 15 base classes and 5 novel classes. We follow \cite{kang2019few} and use the same base/novel splits and support images. We report AP50 results under shots 1, 2, 3, 5, and 10.

More implementation details are included in the Supplementary file.

\subsection{Ablation Study}
\label{ablation}

\begin{figure}[t]
\begin{center}
\includegraphics[scale=0.26]{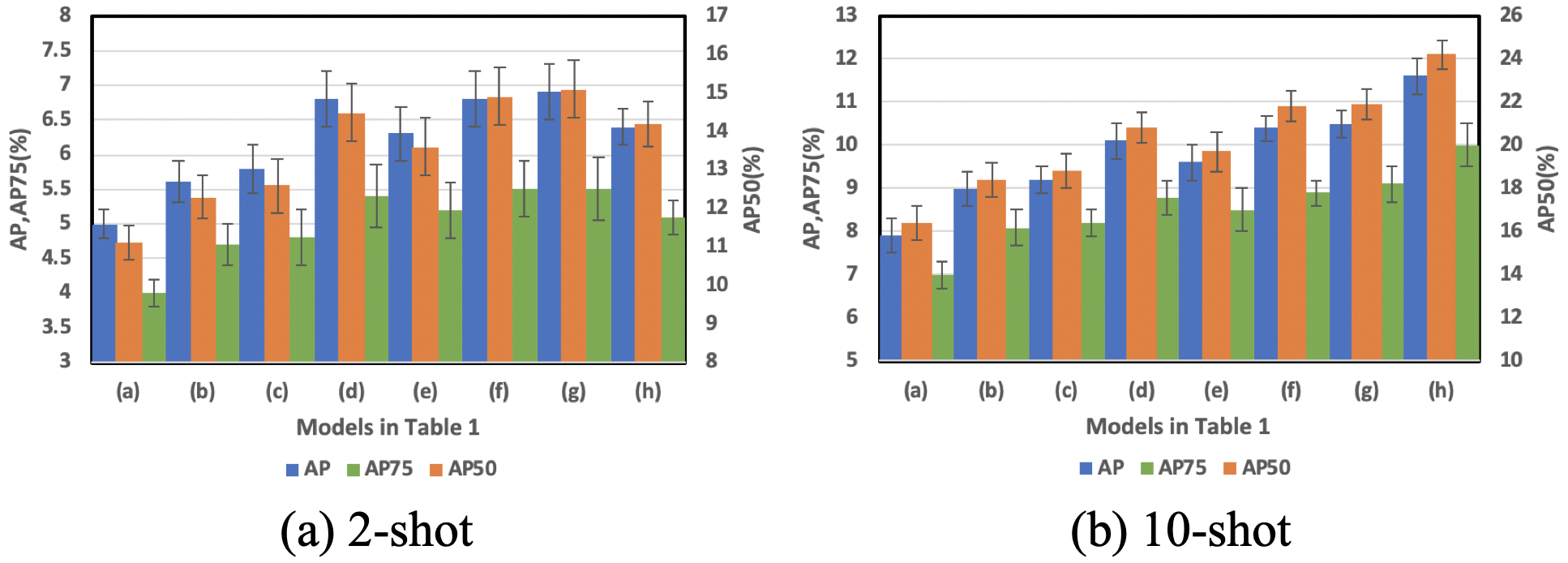}
\end{center}
\caption{The average accuracy and standard deviation results of models in Table \ref{tab:three_edges} over 10 runs.}
\label{tab:multi_runs}
\end{figure}

\textbf{How do the graph convolutional layers help for FSOD?} As shown in the Table \ref{tab:three_edges} and Fig.\ \ref{tab:multi_runs}, we analyze the impact of each component in our model. We first verify the effectiveness of the GCN layers. 
To this end, we replace the GCN layers with fully-connected layers (MLP). Specifically, if we only consider a one-layer GCN, Eq.\ \ref{GCN} becomes $\Mat{Y}=\Mat{AXW}$, where $\Mat{A}$ is the adjacency matrix, and $\Mat{W}$ is the learnable parameter. Notice that the MLP baseline shares the same structure as GCN except that we remove the adjacency matrix $\Mat{A}$. In other words, the MLP can be formulated as $\Mat{Y}=\Mat{XW}$. Compared with the GCN layer, the MLP baseline only uses self-connected edges in the graph, and as a result, each node updates its features independently. Comparing the MLP baseline in Table \ref{tab:three_edges} (b) with the vanilla baseline model in Table \ref{tab:three_edges} (a) and our heterogeneous GCNs in Table \ref{tab:three_edges} (g), we can conclude that the additional learnable modules are useful, and the message passing among different nodes in the graph is crucial for the final performance. 

\textbf{How do the three types of edges help for FSOD?} We then analyze the roles of the three types of edges in our heterogeneous graph. Firstly, we experiment on using only one type of edges as shown in Table \ref{tab:three_edges} (c-e). We notice that all three types of edges can improve the baseline model's performance in Table \ref{tab:three_edges} (b). This demonstrates the effectiveness of multi-class modeling, class-proposal mutual adaptation, and learning context-aware proposal features in our model. Among all three types of edges, we observe that the class-proposal edge is the most important. This is because the objective of FSOD is to calculate the similarity score between the proposal feature and class prototype, thereby preferring that both sides adapt to each other. By further adding proposal-proposal edges, we obtain the full Intra-Class Subgraph, which, as shown in Table \ref{tab:three_edges} (f), further improves the performance owing to context-aware proposal features. Finally, our full model, shown in Table \ref{tab:three_edges} (g), achieves the best results after introducing the Inter-Class Subgraph.

\textbf{The effectiveness of meta-learning and fine-tuning.} We show the comparison results between meta-training-only and fine-tuning in Table \ref{tab:three_edges} (g) and (h). We find that fine-tuning improves the performance in 10/30 shot settings. However, when examples are extremely scarce, e.g., 2-shot as in Table \ref{tab:three_edges}, the performance hardly improves as fine-tuning tends to over-fit with small samples. This demonstrates the strong generalization ability of our meta-learning-based model, and fine-tuning needs large number of samples to perform well. 

\textbf{The roles of local and global context in the proposal-proposal edge.} We show the ablation study of using local and global context in Table \ref{tab:proposal_proposal}. We find that both local and global contextual information contributes to the model's performance. Local context can provide missing features and help refine bounding boxes, especially when proposals are not accurate. Global scene-context, on the other hand, can provide complementary information from the global view. Using both context produces the best results.

\textbf{The efficacy of the bidirectional class-proposal edge.} We show in Table \ref{tab:class_proposal} the results of using bidirectional class-proposal edges against using solely uni-directional edges. We observe that the model with mutual adaptation achieves better results than with any of the two types of uni-directional edges.

\textbf{The comparison between the class-proposal edge and non-local attention in \cite{hsieh2019one}.} We compare our proposed method with \textit{Hsieh et al.}'s work \cite{hsieh2019one} in Table \ref{tab:class_proposal}. \textit{Hsieh et al.}\ \cite{hsieh2019one} proposes to use non-local attention (a.k.a co-attention) between the query image and the support image for feature enhancement. To compare its performance with our class-proposal edge, we use the official codebase of \cite{hsieh2019one} and perform training/testing in our FSOD pipeline. Our model outperforms \textit{Hsieh et al.}'s method \cite{hsieh2019one} significantly. The main difference is that in \cite{hsieh2019one}, each `pixel' in the feature maps is regarded as a basic unit for co-attention. In contrast, the category-specific proposal is used as the basic node in our model. Compared with the `pixels' in the feature maps used in \cite{hsieh2019one}, the category-specific proposals could provide richer semantics related to the target object in the query image, and therefore are more suitable for mutual adaptation.

\textbf{The advantage of using base class memory in the Inter-Class Subgraph.} We show in Table \ref{tab:base_class_memory} the results of using different numbers of base classes in our Inter-Class Subgraph. If we only use novel classes, the model gains little advantage from multi-class modeling. However, the Inter-Class Subgraph sees massive improvement when introduced with all base-class prototypes, which could enhance novel-class prototypes with more robust features from base classes. In practice, we deploy all 60 base classes by default in our Inter-Class Subgraph.

\begin{table*}[t]
    \centering
    \footnotesize
    \caption{Ablation study on each component of the our model.}
    \adjustbox{width=\linewidth}{
    \begin{tabular}{c|ccc|ccc|ccc|ccc}
    \toprule
    &\multirow{2}{*}{Class-class} & \multirow{2}{*}{Class-proposal} & \multirow{2}{*}{Proposal-proposal}
    & \multicolumn{3}{c|}{2-shot} & \multicolumn{3}{c}{10-shot} & \multicolumn{3}{c}{30-shot} \\
    & & & & AP & AP50 & AP75 & AP & AP50 & AP75 & AP & AP50 & AP75 \\ \midrule
    \multicolumn{13}{c}{\textbf{Meta-training the model on base classes, and meta-testing on novel classes}} \\ \midrule
    (a) & \multicolumn{3}{c|}{w/o heterogeneous GCNs} & 5.4 & 11.6 & 4.6 &   7.6 & 15.4 & 6.8  & 8.9 & 17.8 & 8.0 \\ 
    (b) & \multicolumn{3}{c|}{Using MLP instead of GCN layers (only self-connected edges)} & 5.9 & 12.5 & 5.1   & 8.4 & 17.0 & 7.6  & 9.8 & 20.3 & 8.8 \\ \midrule
    (c) & \checkmark & & & 6.3 & 13.3 & 5.5   & 9.0 & 17.7 & 8.1   & 10.6 & 20.9 & 9.6 \\
    (d) & & \checkmark & & 7.6 & 16.0 & 6.5   & 9.8 & 19.7 & 8.8   & 11.2 & 22.8 & 10.1 \\
    (e) & & & \checkmark & 6.7 & 14.0 & 5.8   & 9.3 & 18.5 & 8.3   & 10.8 & 21.5 & 9.7 \\
    (f) & & \checkmark & \checkmark & 7.6 & 16.2 & 6.5   & 10.0 & 20.1 & 8.9   & 11.3 & 23.1 & 10.1 \\
    (g) & \checkmark & \checkmark & \checkmark & \textbf{7.8} & \textbf{16.4} & \textbf{6.6}  & {10.2} & {20.4} & {9.0}   & {11.5} & {23.4} & {10.3} \\ \midrule
    \multicolumn{13}{c}{\textbf{Fine-tuning the model on novel classes, and testing on novel classes}} \\ \midrule
    (h) & \checkmark & \checkmark & \checkmark & 7.6 & 16.1 & 6.2   & \textbf{11.6} & \textbf{23.9} & \textbf{9.8} & \textbf{16.5} & \textbf{31.9} & \textbf{15.5} \\ 
    \bottomrule
    \end{tabular}}
\label{tab:three_edges}
\end{table*}

\begin{table}[t]
    \centering
    \footnotesize
    \caption{Ablation study on the proposal-proposal edge.}
    \adjustbox{width=\linewidth}{
    \begin{tabular}{cc|ccc|ccc}
    \toprule
    \multirow{2}{*}{Local} & \multirow{2}{*}{Global}
    & \multicolumn{3}{c|}{2-shot} & \multicolumn{3}{c}{10-shot}\\
    & & AP & AP50 & AP75 & AP & AP50 & AP75 \\ \midrule
    \checkmark & & 6.6 & 13.8 & 5.7  & 9.0 & 17.9 & 8.1 \\
    & \checkmark & 6.4 & 13.2 & 5.7  & 9.1 & 17.9 & 8.2 \\
    \checkmark & \checkmark & \textbf{6.7} & \textbf{14.0} & \textbf{5.8}  & \textbf{9.3} & \textbf{18.5} & \textbf{8.3} \\ 
    \bottomrule
    \end{tabular}}
\label{tab:proposal_proposal}
\end{table}

\begin{table}[t]
    \centering
    \footnotesize
    \caption{Ablation study on the class-proposal edge.}
    \adjustbox{width=\linewidth}{
    \begin{tabular}{l|ccc|ccc}
    \toprule
    \multirow{2}{*}{Model} & \multicolumn{3}{c|}{2-shot} & \multicolumn{3}{c}{10-shot}\\
    & AP & AP50 & AP75 & AP & AP50 & AP75 \\ \midrule
    class$\rightarrow$proposal & 6.4 & 13.2 & 5.9  & 8.6 & 17.1 & 7.9 \\
    class$\leftarrow$proposal & 7.2 & 15.7 & 5.7  & 9.4 & 19.1 & 8.6 \\ 
    class$\leftrightarrow$proposal & \textbf{7.6} & \textbf{16.0} & \textbf{6.5}  & \textbf{9.8} & \textbf{19.7} & \textbf{8.8} \\ \midrule
    Non-local attention \cite{hsieh2019one} & 6.0 & 12.8 & 5.3  & 8.3 & 17.3 & 7.4 \\
    \bottomrule
    \end{tabular}}
\label{tab:class_proposal}
\end{table}

\begin{table}[t]
    \centering
    \footnotesize
    \caption{Ablation study on the number of GCN layers for the Inter-Class Subgraph. }
    \adjustbox{width=\linewidth}{
    \begin{tabular}{l|ccc|ccc}
    \toprule
    \multirow{2}{*}{\#GCN Layer} & \multicolumn{3}{c|}{2-shot} & \multicolumn{3}{c}{10-shot}\\
    & AP & AP50 & AP75 & AP & AP50 & AP75 \\ \midrule
    1 w/ W & 5.0 & 10.4 & 4.3  & 7.2 & 14.2 & 6.5 \\ \midrule
    1 w/o W & \textbf{6.3} & \textbf{13.3} & \textbf{5.5}  & \textbf{9.0} & \textbf{17.7} & \textbf{8.1} \\
    2 w/o W & 6.1 & 13.0 & 5.3  & 8.8 & 17.6 & 8.0 \\
    3 w/o W & 5.8 & 12.2 & 4.9    & 8.6 & 17.3 & 7.8 \\
    \bottomrule
    \end{tabular}}
\label{tab:GCN_Layer_first}
\end{table}

\begin{table}[t]
    \centering
    \footnotesize
    \caption{Ablation study on the number of GCN layers for the Intra-Class Subgraph. }
    \adjustbox{width=\linewidth}{
    \begin{tabular}{l|ccc|ccc}
    \toprule
    \multirow{2}{*}{\#GCN Layer} & \multicolumn{3}{c|}{2-shot} & \multicolumn{3}{c}{10-shot}\\
    & AP & AP50 & AP75 & AP & AP50 & AP75 \\ \midrule
    1 w/o W & 7.2 & 15.6 & 6.0  & 9.4 & 19.3 & 8.4 \\ \midrule
    1 w/ W & \textbf{7.6} & \textbf{16.2} & \textbf{6.5}  & \textbf{10.0} & \textbf{20.1} & \textbf{8.9} \\
    2 w/ W & 7.4 & 15.9 & 6.3  & 9.8 & 19.7 & 8.8 \\
    3 w/ W & 6.9 & 14.5 & 5.5  & 9.2 & 18.6 & 8.1 \\
    \bottomrule
    \end{tabular}}
\label{tab:GCN_Layer_second}
\end{table}

\begin{table}[t]
    \centering
    \footnotesize
    \caption{Ablation study on base class memory in the Inter-Class Subgraph.}
    \adjustbox{width=\linewidth}{
    \begin{tabular}{l|ccc|ccc}
    \toprule
    \multirow{2}{*}{\#Base Classes} & \multicolumn{3}{c|}{2-shot} & \multicolumn{3}{c}{10-shot}\\
    & AP & AP50 & AP75 & AP & AP50 & AP75 \\ \midrule
    0 & 6.0 & 12.8 & 5.1  & 8.5 & 17.0 & 7.7 \\
    20 & 6.1 & 13.1 & 5.2  & 8.6 & 17.3 & 7.7 \\
    40 & 6.3 & \textbf{13.4} & 5.4  & 8.9 & 17.5 & 8.0 \\
    60 & \textbf{6.3} & 13.3 & \textbf{5.5}  & \textbf{9.0} & \textbf{17.7} & \textbf{8.1} \\
    \bottomrule
    \end{tabular}}
\label{tab:base_class_memory}
\end{table}

\textbf{The analysis on different numbers of GCN layers in the Inter-Class and the Intra-Class Subgraphs.} We show in Table \ref{tab:GCN_Layer_first} and \ref{tab:GCN_Layer_second} the results of applying different numbers of GCN layers in the two Subgraphs.
\textbf{(1)} We first emphasize that proposal and class nodes should go through the same number of transformation layers in a siamese network before the final pairwise matching. 
If we learn $W$ for the inter-class subgraph, the few-shot classes would have one more learnable layer than the proposals, and we show in Table \ref{tab:GCN_Layer_first} that the performance is worse. But for intra-class subgraphs, the GCN layers are applied to both proposals and few-shot classes. As shown in Table \ref{tab:GCN_Layer_second}, learning $W$ does not violate the siamese structure, and could help improve performance.
\textbf{(2)} As shown in both Table \ref{tab:GCN_Layer_first} and \ref{tab:GCN_Layer_second}, using one GCN layer is sufficient in both subgraphs as we already connect edges to all neighbors that a node needs in our model. 
Using more GCN layers are not helpful due to the over-smoothing problem \cite{li2018deeper} in GCNs.

\begin{table*}[ht]
\centering
\footnotesize
\setlength{\tabcolsep}{0.4em}
\caption{{Few-shot object detection performance (AP50) on the PASCAL VOC dataset.} $^{\dag}$We re-evaluate the methods following the standard procedure in \cite{kang2019few,wang2020few}.
Our approach with only meta-learning could achieve competitive results compared with other methods using fine-tuning, especially on extreme few-shot settings. After fine-tuning, our model outperforms other methods in almost all settings.
}
\adjustbox{width=\linewidth}{
\begin{tabular}{l|c|c|ccccc|ccccc|ccccc}
\toprule
\multirow{2}{*}{Method / Shot} & \multirow{2}{*}{Venue} & \multirow{2}{*}{Backbone} & \multicolumn{5}{c|}{Novel Set 1} & \multicolumn{5}{c|}{Novel Set 2} & \multicolumn{5}{c}{Novel Set 3} \\ 
&  & & 1     & 2     & 3    & 5    & 10   & 1     & 2     & 3    & 5    & 10   & 1     & 2     & 3    & 5    & 10   \\ \midrule
\multicolumn{18}{c}{\textbf{Meta-training the model on base classes, and meta-testing on novel classes}} \\ \midrule
Fan et al.\ \cite{fan2020few}$^{\dag}$ & CVPR 2020 & ResNet-101 & 32.4 & 22.1 & 23.1 & 31.7 & 35.7    & 14.8 & 18.1 & 24.4 & 18.6 & 19.5 &    25.8 & 20.9 & 23.9 & 27.8 & 29.0 \\
QA-FewDet (Ours) & This work & ResNet-101 & \textbf{41.0} & \textbf{33.2} & \textbf{35.3} & \textbf{47.5} & \textbf{52.0}   & \textbf{23.5} & \textbf{29.4} & \textbf{37.9} & \textbf{35.9} & \textbf{37.1}   & \textbf{33.2} & \textbf{29.4} & \textbf{37.6} & \textbf{39.8} & \textbf{41.5} \\ \midrule
\multicolumn{18}{c}{\textbf{Fine-tuning the model on novel classes, and testing on novel classes}} \\ \midrule
FSRW~\cite{kang2019few}  & ICCV 2019 & YOLOv2 & 14.8  & 15.5  & 26.7 & 33.9 & 47.2 & 15.7  & 15.3  & 22.7 & 30.1 & 40.5 & 21.3  & 25.6  & 28.4 & 42.8 & 45.9 \\ 
MetaDet~\cite{wang2019meta} & ICCV 2019 & VGG16 & 18.9 & 20.6 & 30.2 & 36.8 & 49.6 & 21.8 & 23.1 & 27.8 & 31.7 & 43.0 & 20.6 & 23.9 & 29.4 & 43.9 & 44.1 \\ 
Meta R-CNN~\cite{yan2019meta} & ICCV 2019 & ResNet-101 & 19.9 & 25.5 & 35.0 & 45.7 & 51.5 & 10.4 & 19.4 & 29.6 & 34.8 & 45.4 & 14.3 & 18.2 & 27.5 & 41.2 & 48.1 \\ 
TFA w/ fc \cite{wang2020few} & ICML 2020 & ResNet-101 & {36.8} & {29.1} & {43.6} & {55.7} & {57.0} & {18.2} & {29.0} & {33.4} & {35.5} & {39.0} & {27.7} & {33.6} & {42.5} & {48.7} & {50.2}\\
TFA w/ cos \cite{wang2020few} & ICML 2020 & ResNet-101 & 39.8 & 36.1 & 44.7 & 55.7 & 56.0 & 23.5 & 26.9 & 34.1 & 35.1 & 39.1 & 30.8 & 34.8 & 42.8 & 49.5 & 49.8 \\ 
Xiao et al.\ \cite{xiao2020few} & ECCV 2020 & ResNet-101 & 24.2 & 35.3 &  42.2 &  49.1 &  57.4 & 21.6 & 24.6 &  31.9 &  37.0 &  45.7 & 21.2 &  30.0 &  37.2 &  43.8 &  49.6 \\
MPSR \cite{wu2020multi} & ECCV 2020 & ResNet-101 & 41.7 & 42.5 & 51.4 & 55.2 & 61.8 & 24.4 & 29.3 & 39.2 & 39.9 & 47.8 & \textbf{35.6} & 41.8 & 42.3 & 48.0 & 49.7 \\ 
Fan et al.\ \cite{fan2020few} $^{\dag}$ & CVPR 2020 & ResNet-101 & 37.8 & 43.6 & 51.6 & 56.5 & 58.6    & 22.5 & 30.6 & 40.7 & 43.1 & 47.6    & 31.0 & 37.9 & 43.7 & 51.3 & 49.8 \\
QA-FewDet (Ours) & This work & ResNet-101 & \textbf{42.4} & \textbf{51.9} & \textbf{55.7} & \textbf{62.6} & \textbf{63.4} & \textbf{25.9} & \textbf{37.8} & \textbf{46.6} & \textbf{48.9} & \textbf{51.1} & 35.2 & \textbf{42.9} & \textbf{47.8} & \textbf{54.8} & \textbf{53.5} \\
\bottomrule
\end{tabular}}
\label{tab:main_voc}
\end{table*}

\begin{table*}[ht]
\centering
\footnotesize
\setlength{\tabcolsep}{0.4em}
\caption{Few-shot object detection performance on the MSCOCO dataset. 
$^{\dag}$We re-evaluate the methods following the standard procedure in \cite{kang2019few,wang2020few}. 
$^{\ddag}$The authors report these results at https://github.com/YoungXIAO13/FewShotDetection. Our method consistently outperforms the state-of-the-art methods in most of the shots and metrics. \vspace{1mm}}
\adjustbox{width=\linewidth}{
\begin{tabular}{l|ccc|ccc|ccc|ccc|ccc|ccc}
\toprule
&\multicolumn{3}{c|}{1-shot} & \multicolumn{3}{c|}{2-shot} & \multicolumn{3}{c|}{3-shot} &\multicolumn{3}{c|}{5-shot} & \multicolumn{3}{c|}{10-shot} & \multicolumn{3}{c}{30-shot} \\
Method & AP & AP50 & AP75 & AP & AP50 & AP75 & AP & AP50 & AP75 & AP & AP50 & AP75 & AP & AP50 & AP75 & AP & AP50 & AP75 \\ \midrule
\multicolumn{19}{c}{\textbf{Meta-training the model on base classes, and meta-testing on novel classes}} \\ \midrule
Fan et al.\ \cite{fan2020few}$^{\dag}$ & 4.0 & 8.5 & 3.5  & 5.4 & 11.6 & 4.6    & 5.9 & 12.5 & 5.0   & 6.9 & 14.3 & 6.0    & 7.6 & 15.4 & 6.8  & 8.9 & 17.8 & 8.0 \\
QA-FewDet (Ours) &  \textbf{5.1} & \textbf{10.5} & \textbf{4.5}    & \textbf{7.8} & \textbf{16.4} & \textbf{6.6}    & \textbf{8.6} & \textbf{17.7} & \textbf{7.5}    & \textbf{9.5} & \textbf{19.3} & \textbf{8.5}  & \textbf{10.2} & \textbf{20.4} & \textbf{9.0}   & \textbf{11.5} & \textbf{23.4} & \textbf{10.3} \\ \midrule 
\multicolumn{18}{c}{\textbf{Fine-tuning the model on novel classes, and testing on novel classes}} \\ \midrule
FSRW\small{~\cite{kang2019few}}  & {--} & {--} & {--} & {--} & {--} & {--} & {--} & {--} & {--} &\;{--} & {--} & {--} & 5.6 & 12.3 & 4.6 & 9.1 & 19.0 & 7.6 \\ 
MetaDet\small{~\cite{wang2019meta}} & {--} & {--} & {--} & {--} & {--} & {--} & {--} & {--} & {--} &\;{--} & {--} & {--} & 7.1 & 14.6 & 6.1 & 11.3 & 21.7 & 8.1 \\
Meta R-CNN \cite{yan2019meta} & {--} & {--} & {--} & {--} & {--} & {--} & {--} & {--} & {--} &\;{--} & {--} & {--} & {8.7} & 19.1 & {6.6} & {12.4} & 25.3 & {10.8} \\
TFA w/ fc \cite{wang2020few} & 2.9 & 5.7 & 2.8   & 4.3 & 8.5 & 4.1    & 6.7 & 12.6 & 6.6   & 8.4 & 16.0 & 8.4    & 10.0 & 19.2 & 9.2    & 13.4 & 24.7 & 13.2 \\
TFA w/ cos \cite{wang2020few} & 3.4 & 5.8 & 3.8   & 4.6 & 8.3 & 4.8    & 6.6 & 12.1 & 6.5   & 8.3 & 15.3 & 8.0     & 10.0 & 19.1 & 9.3    & 13.7& 24.9 & 13.4 \\
Xiao et al.\ \cite{xiao2020few}$^{\ddag}$ & 3.2 & 8.9 & 1.4  & 4.9 & 13.3 & 2.3  & 6.7 & \textbf{18.6} & 2.9  & 8.1 & 20.1 & 4.4 & 10.7 & \textbf{25.6} & 6.5 & 15.9 & 31.7 & 15.1 \\
MPSR \cite{wu2020multi} $^{\dag}$        & {2.3} & {4.1} & {2.3}    & {3.5} & {6.3} & {3.4}    & {5.2} & {9.5} & {5.1}   &{6.7} & {12.6} & {6.4}    & {9.8} & {17.9} & {9.7}   & {14.1} & {25.4} & {14.2} \\
Fan et al.\ \cite{fan2020few}$^{\dag}$ & 4.2 & 9.1 & 3.0   & 5.6 & 14.0 & 3.9   & 6.6 & 15.9 & 4.9   & 8.0 & 18.5 & 6.3   & 9.6 & 20.7 & 7.7    & 13.5 & 28.5 & 11.7 \\
QA-FewDet (Ours) & \textbf{4.9} & \textbf{10.3} & \textbf{4.4}  & \textbf{7.6} & \textbf{16.1} & \textbf{6.2}    & \textbf{8.4} & 18.0 & \textbf{7.3}   & \textbf{9.7} & \textbf{20.3} & \textbf{8.6}   & \textbf{11.6} & 23.9 & \textbf{9.8} & \textbf{16.5} & \textbf{31.9} & \textbf{15.5} \\
\bottomrule
\end{tabular}}
\label{tab:main_coco}
\vspace{-2mm}
\end{table*}

\subsection{Comparison with State-of-the-arts}

\label{compare_with_SOTA}

As shown in Table \ref{tab:main_voc} and \ref{tab:main_coco}, we compare our QA-FewDet with the STOAs on PASCAL VOC and MSCOCO FSOD benchmarks. We draw the following three conclusion: \textbf{(1)} Our final model significantly outperforms previous STOAs by more than $4.0\%$ on AP50 in most of the shots and metrics of the PASCAL VOC. We achieve similar improvement on the MSCOCO. \textbf{(2)} Fine-tuning does not help too much in extreme few-shot settings because it is prone to over-fitting with very few samples (e.g., 1-shot in the PASCAL VOC, and 1/2/3-shot in the MSCOCO.), but could help in larger shot settings. \textbf{(3)} Our meta-learning-only model improves significantly compared with the strong baseline model \cite{fan2020few}, and outperforms or at least attains comparable results compared with other SOTAs using fine-tuning on 1/2-shot in the PASCAL VOC and on 1/2/3/5/10-shot in the MSCOCO.

\section{Conclusion}
In this paper, we introduce a novel heterogeneous GCNs that consider multi-relations among the proposal and class nodes for FSOD. The Inter-Class Subgraph enhances novel-class prototype representation via modeling multi-class relations. The Intra-Class Subgraph provides query-adaptive class prototypes and context-aware proposal features to facilitate pairwise matching. Our experiments show that our model, QA-FewDet, with only meta-learning, can outperform or achieve competitive results especially on extreme few-shot settings. After fine-tuning, our model outperforms current SOTAs by a large margin across various settings.

\textbf{Acknowledgment}
This research is based upon work supported by the Intelligence
Advanced Research Projects Activity (IARPA) via Department
of Interior/Interior Business Center (DOI/IBC) contract number
D17PC00345. The U.S. Government is authorized to reproduce and
distribute reprints for Governmental purposes not withstanding any
copyright annotation theron. Disclaimer: The views and conclusions
contained herein are those of the authors and should not be
interpreted as necessarily representing the official policies or
endorsements, either expressed or implied of IARPA, DOI/IBC or
the U.S. Government.

{\small
\bibliographystyle{ieee_fullname}
\bibliography{QA-FewDet}

\begin{thebibliography}{10}\itemsep=-1pt

\bibitem{chen2020hierarchical}
Jintai Chen, Biwen Lei, Qingyu Song, Haochao Ying, Danny~Z Chen, and Jian Wu.
\newblock A hierarchical graph network for 3d object detection on point clouds.
\newblock In {\em Proceedings of the IEEE/CVF Conference on Computer Vision and
  Pattern Recognition}, pages 392--401, 2020.

\bibitem{everingham2010pascal}
Mark Everingham, Luc Van~Gool, Christopher~KI Williams, John Winn, and Andrew
  Zisserman.
\newblock The pascal visual object classes (voc) challenge.
\newblock {\em International journal of computer vision}, 88(2):303--338, 2010.

\bibitem{fan2020few}
Qi Fan, Wei Zhuo, Chi-Keung Tang, and Yu-Wing Tai.
\newblock Few-shot object detection with attention-rpn and multi-relation
  detector.
\newblock In {\em Proceedings of the IEEE/CVF Conference on Computer Vision and
  Pattern Recognition}, pages 4013--4022, 2020.

\bibitem{finn2017model}
Chelsea Finn, Pieter Abbeel, and Sergey Levine.
\newblock Model-agnostic meta-learning for fast adaptation of deep networks.
\newblock In {\em International Conference on Machine Learning}, pages
  1126--1135, 2017.

\bibitem{gidaris2018dynamic}
Spyros Gidaris and Nikos Komodakis.
\newblock Dynamic few-shot visual learning without forgetting.
\newblock In {\em Proceedings of the IEEE Conference on Computer Vision and
  Pattern Recognition}, pages 4367--4375, 2018.

\bibitem{Fast_R-CNN}
Ross Girshick.
\newblock Fast r-cnn.
\newblock In {\em Proceedings of the IEEE international conference on computer
  vision}, pages 1440--1448, 2015.

\bibitem{han2021meta}
Guangxing Han, Shiyuan Huang, Jiawei Ma, Yicheng He, and Shih-Fu Chang.
\newblock Meta faster r-cnn: Towards accurate few-shot object detection with
  attentive feature alignment.
\newblock In {\em AAAI}, 2022.

\bibitem{R_RPN}
Guangxing Han, Xuan Zhang, and Chongrong Li.
\newblock Revisiting faster r-cnn: A deeper look at region proposal network.
\newblock In {\em ICONIP}, pages 14--24, 2017.

\bibitem{SSD_TDR}
Guangxing Han, Xuan Zhang, and Chongrong Li.
\newblock Single shot object detection with top-down refinement.
\newblock In {\em ICIP}, pages 3360--3364, 2017.

\bibitem{han2018semi}
Guangxing Han, Xuan Zhang, and Chongrong Li.
\newblock Semi-supervised dff: Decoupling detection and feature flow for video
  object detectors.
\newblock In {\em Proceedings of the 26th ACM international conference on
  Multimedia}, pages 1811--1819, 2018.

\bibitem{he2017mask}
Kaiming He, Georgia Gkioxari, Piotr Doll{\'a}r, and Ross Girshick.
\newblock Mask r-cnn.
\newblock In {\em Proceedings of the IEEE international conference on computer
  vision}, pages 2961--2969, 2017.

\bibitem{he2016deep}
Kaiming He, Xiangyu Zhang, Shaoqing Ren, and Jian Sun.
\newblock Deep residual learning for image recognition.
\newblock In {\em Proceedings of the IEEE conference on computer vision and
  pattern recognition}, pages 770--778, 2016.

\bibitem{hinton2015distilling}
Geoffrey Hinton, Oriol Vinyals, and Jeffrey Dean.
\newblock Distilling the knowledge in a neural network.
\newblock In {\em NIPS Deep Learning and Representation Learning Workshop},
  2014.

\bibitem{hsieh2019one}
Ting-I Hsieh, Yi-Chen Lo, Hwann-Tzong Chen, and Tyng-Luh Liu.
\newblock One-shot object detection with co-attention and co-excitation.
\newblock In {\em Advances in Neural Information Processing Systems}, pages
  2725--2734, 2019.

\bibitem{hu2018relation}
Han Hu, Jiayuan Gu, Zheng Zhang, Jifeng Dai, and Yichen Wei.
\newblock Relation networks for object detection.
\newblock In {\em Proceedings of the IEEE Conference on Computer Vision and
  Pattern Recognition}, pages 3588--3597, 2018.

\bibitem{kang2019few}
Bingyi Kang, Zhuang Liu, Xin Wang, Fisher Yu, Jiashi Feng, and Trevor Darrell.
\newblock Few-shot object detection via feature reweighting.
\newblock In {\em Proceedings of the IEEE International Conference on Computer
  Vision}, pages 8420--8429, 2019.

\bibitem{kang2019decoupling}
Bingyi Kang, Saining Xie, Marcus Rohrbach, Zhicheng Yan, Albert Gordo, Jiashi
  Feng, and Yannis Kalantidis.
\newblock Decoupling representation and classifier for long-tailed recognition.
\newblock In {\em Eighth International Conference on Learning Representations
  (ICLR)}, 2020.

\bibitem{karlinsky2019repmet}
Leonid Karlinsky, Joseph Shtok, Sivan Harary, Eli Schwartz, Amit Aides, Rogerio
  Feris, Raja Giryes, and Alex~M Bronstein.
\newblock Repmet: Representative-based metric learning for classification and
  few-shot object detection.
\newblock In {\em Proceedings of the IEEE Conference on Computer Vision and
  Pattern Recognition}, pages 5197--5206, 2019.

\bibitem{kipf2016semi}
Thomas~N. Kipf and Max Welling.
\newblock Semi-supervised classification with graph convolutional networks.
\newblock In {\em International Conference on Learning Representations (ICLR)},
  2017.

\bibitem{li2018deeper}
Qimai Li, Zhichao Han, and Xiao-Ming Wu.
\newblock Deeper insights into graph convolutional networks for semi-supervised
  learning.
\newblock In {\em Proceedings of the AAAI Conference on Artificial
  Intelligence}, volume~32, 2018.

\bibitem{lin2017focal}
Tsung-Yi Lin, Priya Goyal, Ross Girshick, Kaiming He, and Piotr Doll{\'a}r.
\newblock Focal loss for dense object detection.
\newblock In {\em Proceedings of the IEEE international conference on computer
  vision}, pages 2980--2988, 2017.

\bibitem{lin2014microsoft}
Tsung-Yi Lin, Michael Maire, Serge Belongie, James Hays, Pietro Perona, Deva
  Ramanan, Piotr Doll{\'a}r, and C~Lawrence Zitnick.
\newblock Microsoft coco: Common objects in context.
\newblock In {\em European conference on computer vision}, pages 740--755.
  Springer, 2014.

\bibitem{liu2016ssd}
Wei Liu, Dragomir Anguelov, Dumitru Erhan, Christian Szegedy, Scott Reed,
  Cheng-Yang Fu, and Alexander~C Berg.
\newblock Ssd: Single shot multibox detector.
\newblock In {\em European conference on computer vision}, pages 21--37.
  Springer, 2016.

\bibitem{liu2018structure}
Yong Liu, Ruiping Wang, Shiguang Shan, and Xilin Chen.
\newblock Structure inference net: Object detection using scene-level context
  and instance-level relationships.
\newblock In {\em Proceedings of the IEEE conference on computer vision and
  pattern recognition}, pages 6985--6994, 2018.

\bibitem{ma2021PAL}
Jiawei Ma, Hanchen Xie, Guangxing Han, Shih-Fu Chang, Aram Galstyan, and Wael
  Abd-Almageed.
\newblock Partner-assisted learning for few-shot image classification.
\newblock {\em Proceedings of the IEEE International Conference on Computer
  Vision}, 2021.

\bibitem{mi2020hierarchical}
Li Mi and Zhenzhong Chen.
\newblock Hierarchical graph attention network for visual relationship
  detection.
\newblock In {\em Proceedings of the IEEE/CVF Conference on Computer Vision and
  Pattern Recognition}, pages 13886--13895, 2020.

\bibitem{nawhal2021activity}
Megha Nawhal and Greg Mori.
\newblock Activity graph transformer for temporal action localization.
\newblock {\em arXiv preprint arXiv:2101.08540}, 2021.

\bibitem{osokin2020os2d}
Anton Osokin, Denis Sumin, and Vasily Lomakin.
\newblock Os2d: One-stage one-shot object detection by matching anchor
  features.
\newblock In {\em European Conference on Computer Vision}, 2020.

\bibitem{perez2020incremental}
Juan-Manuel Perez-Rua, Xiatian Zhu, Timothy~M Hospedales, and Tao Xiang.
\newblock Incremental few-shot object detection.
\newblock In {\em Proceedings of the IEEE/CVF Conference on Computer Vision and
  Pattern Recognition}, pages 13846--13855, 2020.

\bibitem{redmon2016you}
Joseph Redmon, Santosh Divvala, Ross Girshick, and Ali Farhadi.
\newblock You only look once: Unified, real-time object detection.
\newblock In {\em Proceedings of the IEEE conference on computer vision and
  pattern recognition}, pages 779--788, 2016.

\bibitem{ren2015faster}
Shaoqing Ren, Kaiming He, Ross Girshick, and Jian Sun.
\newblock Faster r-cnn: Towards real-time object detection with region proposal
  networks.
\newblock In {\em Advances in neural information processing systems}, pages
  91--99, 2015.

\bibitem{snell2017prototypical}
Jake Snell, Kevin Swersky, and Richard Zemel.
\newblock Prototypical networks for few-shot learning.
\newblock In {\em Advances in neural information processing systems}, pages
  4077--4087, 2017.

\bibitem{sung2018learning}
Flood Sung, Yongxin Yang, Li Zhang, Tao Xiang, Philip~HS Torr, and Timothy~M
  Hospedales.
\newblock Learning to compare: Relation network for few-shot learning.
\newblock In {\em Proceedings of the IEEE Conference on Computer Vision and
  Pattern Recognition}, pages 1199--1208, 2018.

\bibitem{tian2019fcos}
Zhi Tian, Chunhua Shen, Hao Chen, and Tong He.
\newblock Fcos: Fully convolutional one-stage object detection.
\newblock In {\em Proceedings of the IEEE international conference on computer
  vision}, pages 9627--9636, 2019.

\bibitem{velivckovic2017graph}
Petar Veli{\v{c}}kovi{\'c}, Guillem Cucurull, Arantxa Casanova, Adriana Romero,
  Pietro Lio, and Yoshua Bengio.
\newblock Graph attention networks.
\newblock In {\em International Conference on Learning Representations (ICLR)},
  2018.

\bibitem{vinyals2016matching}
Oriol Vinyals, Charles Blundell, Timothy Lillicrap, Daan Wierstra, et~al.
\newblock Matching networks for one shot learning.
\newblock In {\em Advances in neural information processing systems}, pages
  3630--3638, 2016.

\bibitem{wang2020few}
Xin Wang, Thomas~E. Huang, Trevor Darrell, Joseph~E Gonzalez, and Fisher Yu.
\newblock Frustratingly simple few-shot object detection.
\newblock In {\em International Conference on Machine Learning (ICML)}, July
  2020.

\bibitem{wang2018zero}
Xiaolong Wang, Yufei Ye, and Abhinav Gupta.
\newblock Zero-shot recognition via semantic embeddings and knowledge graphs.
\newblock In {\em Proceedings of the IEEE conference on computer vision and
  pattern recognition}, pages 6857--6866, 2018.

\bibitem{wang2019meta}
Yu-Xiong Wang, Deva Ramanan, and Martial Hebert.
\newblock Meta-learning to detect rare objects.
\newblock In {\em Proceedings of the IEEE International Conference on Computer
  Vision}, pages 9925--9934, 2019.

\bibitem{wu2020multi}
Jiaxi Wu, Songtao Liu, Di Huang, and Yunhong Wang.
\newblock Multi-scale positive sample refinement for few-shot object detection.
\newblock In {\em European Conference on Computer Vision}, pages 456--472.
  Springer, 2020.

\bibitem{xiao2020few}
Yang Xiao and Renaud Marlet.
\newblock Few-shot object detection and viewpoint estimation for objects in the
  wild.
\newblock In {\em European Conference on Computer Vision}, 2020.

\bibitem{yan2018spatial}
Sijie Yan, Yuanjun Xiong, and Dahua Lin.
\newblock Spatial temporal graph convolutional networks for skeleton-based
  action recognition.
\newblock In {\em Proceedings of the AAAI conference on artificial
  intelligence}, volume~32, 2018.

\bibitem{yan2019meta}
Xiaopeng Yan, Ziliang Chen, Anni Xu, Xiaoxi Wang, Xiaodan Liang, and Liang Lin.
\newblock Meta r-cnn: Towards general solver for instance-level low-shot
  learning.
\newblock In {\em Proceedings of the IEEE International Conference on Computer
  Vision}, pages 9577--9586, 2019.

\bibitem{yang2020restoring}
Yukuan Yang, Fangyu Wei, Miaojing Shi, and Guoqi Li.
\newblock Restoring negative information in few-shot object detection.
\newblock In {\em Advances in neural information processing systems}, 2020.

\bibitem{zeng2019graph}
Runhao Zeng, Wenbing Huang, Mingkui Tan, Yu Rong, Peilin Zhao, Junzhou Huang,
  and Chuang Gan.
\newblock Graph convolutional networks for temporal action localization.
\newblock In {\em Proceedings of the IEEE/CVF International Conference on
  Computer Vision}, pages 7094--7103, 2019.

\end{thebibliography}
}

\clearpage
\newpage
\appendix

\section*{Appendix}
The supplementary materials are organized as follows. First, we describe the implementation details of our model architecture in Section \ref{Overall_Pipeline}. Then, we describe the implementation details of our training framework in Section \ref{Training_Framework}. We show the visualization of the category-specific proposals in Section \ref{proposals_section}, and the visualization of the pairwise cosine similarity of all classes (80) in MSCOCO in Section \ref{Paiwise_cosine_section}.

\begin{figure*}[t]
\begin{center}
\includegraphics[scale=0.35]{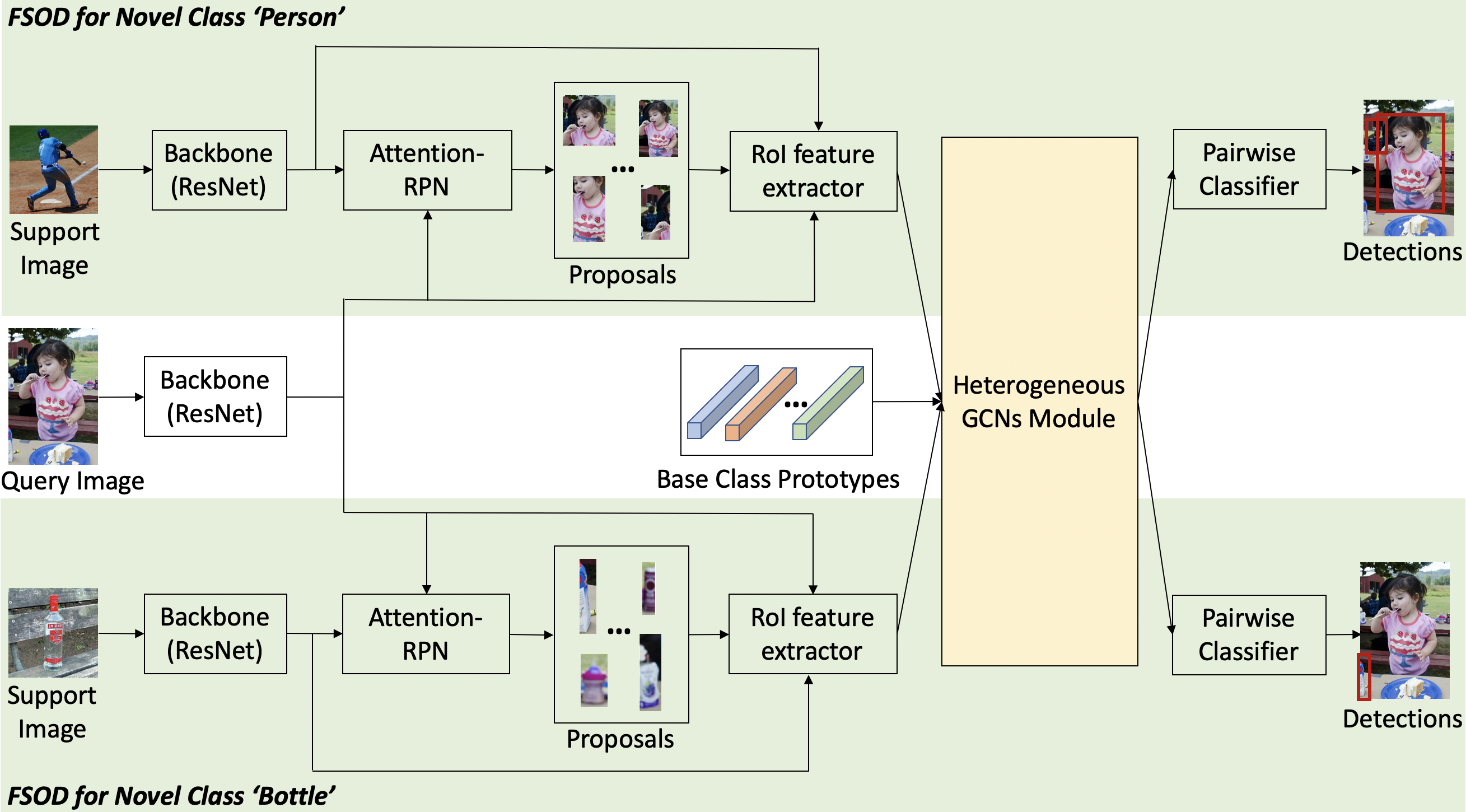}
\end{center}
\caption{Our model architecture. We use two novel classes as an example. The backbone feature extractor, the Attention-RPN \cite{fan2020few}, the RoI feature extractor the and pairwise classifier \cite{fan2020few} are shared among all the branches.}
\label{pipeline}
\vspace{-2mm}
\end{figure*}

\begin{figure}[h]
\begin{center}
\includegraphics[scale=0.25]{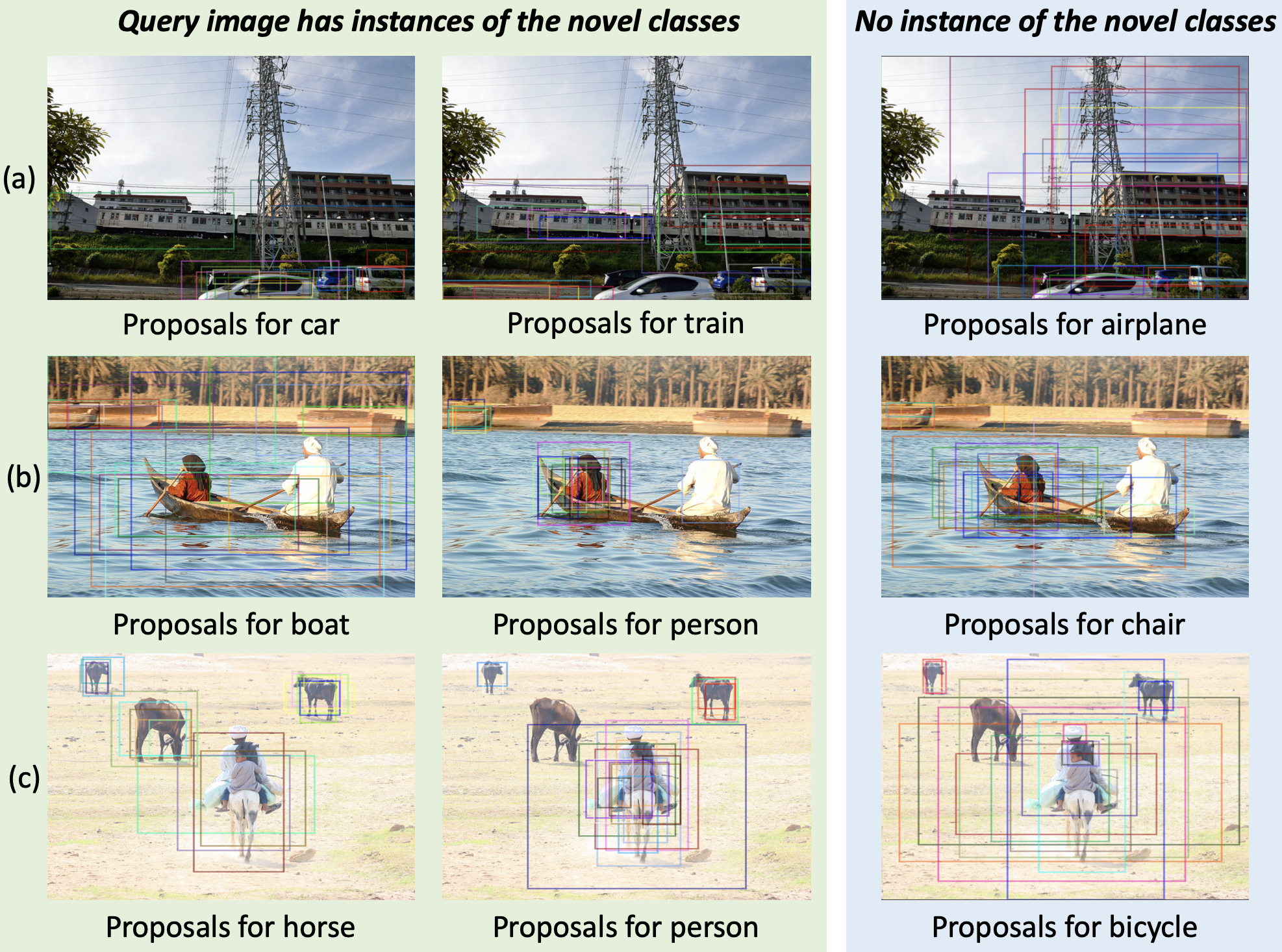}
\end{center}
\caption{Visualization of category-specific proposals generated using Attention-RPN in \cite{fan2020few}. We show both cases of the query images having instances of the target class and not.}
\label{proposals}
\vspace{-2mm}
\end{figure}

\begin{figure*}[h]
\begin{center}
\includegraphics[scale=0.32]{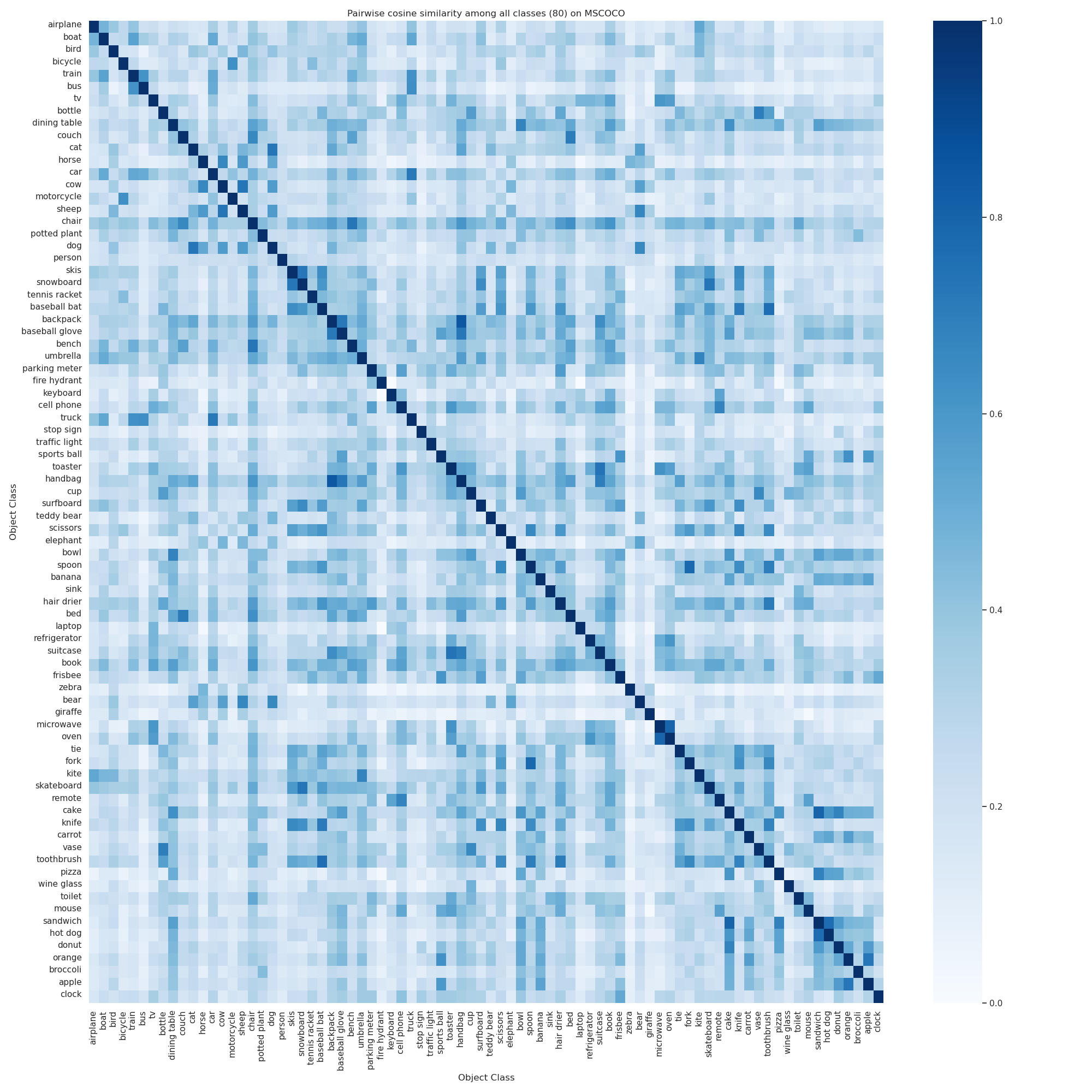}
\end{center}
\caption{Pairwise cosine similarity of all classes (80) in MSCOCO. The first 20 classes are novel classes (from `airplane' to `person'), and the rest are base classes.}
\label{Paiwise_cosine}
\end{figure*}

\section{Our Model Architecture with Implementation Details}
\label{Overall_Pipeline}

We illustrate our model architecture in Fig.\ \ref{pipeline}. Our model is built upon the baseline FSOD model in \cite{fan2020few}. Next we describe each component in our model in details, including the backbone feature extractor, the Attention-RPN \cite{fan2020few}, the RoI feature extractor, the heterogeneous GCNs and the pairwise classifier \cite{fan2020few}.

\subsection{The backbone feature extractor}

We use ResNet-50/101 as our backbone feature extractor by default for fair comparison with other STOAs. Specifically, we use the output of the $res4$ block as the input image features of the detection head. The backbone feature extractor is shared for both query image $I_q$ and support image $I_s$ to extract the features $r(I_q)$ and $r(I_s)$ respectively.

\subsection{The Attention-RPN}

Following \cite{fan2020few}, we use the Attention-RPN module to generate category-specific proposals for each novel class. Formally, for each novel class $c \in \mathcal{C}_{novel}$, we first take the average feature of all support images belonging to that novel class as $r(c)$, and then conduct spatial average pooling to get the spatial-averaged feature $r(c)_{pool} = \frac{1}{H*W}\sum_{h,w} r(c)$, such that $r(c)_{pool}$ captures global representation of class $c$. Then we modulate the query image features using $r(c)_{pool}$, which can highlight important and relevant features in the query image for class $c$,
\begin{equation}
r(I_q)_c = r(I_q) \odot r(c)_{pool}, c \in \mathcal{C}_{novel}
\end{equation}
where $\odot$ represents channel-wise Hadamard product.

After that, we use the original RPN to generate proposals using the modulated query image features, such that the proposals are category-specific. Typically we generate 100 proposals for each novel class by default.

\subsection{The RoI feature extractor}

After generating the proposals in the query image, we use RoI Align and the $res5$ block in ResNet to extract the feature $f(p_i^c)$ for proposal $p_i^c$ from the query image feature $r(I_q)$. The same layers are applied to $r(I_s)$ for each support image, and we take the average feature of all support images belonging to the novel class $c$ as the class prototype $f(c)$.

\subsection{Our Heterogeneous GCNs Module}

We propose the heterogeneous GCNs in this paper to enable efficient message passing among the proposal and class nodes, such that we could learn context-aware proposal features and query-adaptive, multiclass-enhanced prototype representations for each class.

For the Inter-Class Subgraph, we build a fully-connected graph (80 nodes) with all base classes (60 nodes) and novel classes (20 nodes) on the MSCOCO dataset. For each base class, we use 30 support images to extract the prototype representation. We emphasize that \textbf{proposals and prototypes should go through the same number of learnable transformation layers} before the final pairwise classification, to make sure the two features are in the same feature space. As shown in Table \ref{tab:GCN_Layer_first} of the main paper, using GCN layers w/o W is better than that w/ W. We adhere to this principle throughout the model design, and do not use the learnable parameter W in our Inter-Class Subgraph. Meanwhile, the Inter-Class Subgraph is query-agnostic and we perform message passing on it before processing any query image.

The Intra-Class Subgraphs are built upon the query image. For each query image, we build an Intra-Class Subgraph for each novel class. For each Intra-Class Subgraph, we have in total 102 nodes, including 1 class node $c$, 1 global image node $g$ and 100 category-specific proposal nodes $P_c$ of the class $c$. We establish proposal-proposal edges between proposals if the $\mathit{IoU}$ is more than the threshold $\theta=0.7$. The far-away proposals could hardly provide meaningful information. We enrich the proposals with global image features by adding edges to the global image node $g$. For the class-proposal edge, we connect bidirectional edges between the class node and all the 100 proposal nodes. 

As shown in Table \ref{tab:GCN_Layer_first} and \ref{tab:GCN_Layer_second} of the main paper, we only use one GCN layer for both the Inter-Class Subgraph and Intra-Class Subgraphs, because we already connect edges to all neighbors that a node needs in our model. Using more GCN layers are not helpful due to the over-smoothing problem \cite{li2018deeper} in GCNs.

\subsection{The pairwise classifier}

After getting the enhanced features for proposals and classes, we use the multi-relation network in \cite{fan2020few} for pairwise classification. Specifically, for each novel class, we calculate the similarity score between the class prototype and category-specific proposal features, and produce the final detection results of this class using post-processing steps in \cite{ren2015faster}.

\section{Implementation Details of Model Training}
\label{Training_Framework}

\textbf{Meta-learning with Base Classes.} We train our FSOD model on base classes using episodic training. 

For model training on the MSCOCO dataset, we first train the baseline FSOD model as in \cite{fan2020few}. We use the SGD optimizer with an initial learning rate of 0.002, momentum of 0.9, weight decay of 0.0001, and a batch size of 8. The learning rate is divided by 10 after 30,000 iterations. The total number of training iterations is 40,000. Then, we add the proposed heterogeneous GCNs and train the whole model. We use a smaller learning rate of 0.001 in this step, which is divided by 10 after 15,000 iterations. The total number of training iterations is 20,000.

Similarly, for model training on the PASCAL VOC dataset, we use the same hyper-parameters as on the MSCOCO dataset except using fewer training iterations. The training iterations are reduced to half in the two steps. 

\textbf{Fine-tuning with Novel Classes.} We fine-tune the few-shot detection model on novel classes in this step. We use the original novel class images to generate positive and negative proposals of the novel classes for training. 

Similar to meta-learning, we use the SGD optimizer with an initial learning rate of 0.001, momentum of 0.9, weight decay of 0.0001, and a batch size of 8. The difference is that we use a much smaller number of training iterations for fine-tuning. The learning rate is divided by 10 after 2,000 iterations, and the total number of training iterations is 3,000. Meanwhile, the backbone feature extractor is fixed during fine-tuning.

As shown in Table \ref{tab:main_voc} and \ref{tab:main_coco} of the main paper, we can conclude that meta-learning is crucial for extreme few-shot (e.g., 1/2 shot) settings due to the strong generalization ability, and fine-tuning turns out to be more useful for larger shot (e.g., 10/30 shot) settings with more training images.

\section{Visualization of the category-specific proposals by Attention-RPN \cite{fan2020few}}
\label{proposals_section}

As shown in Fig.\ \ref{proposals}, we can find that if the query image contains instances of the target class, the category-specific proposals would be more likely the nearby regions. If not, the proposals could be regions of similar classes or just random regions. Motivated by this observation, we propose to build multiple class-specific Intra-Class Subgraphs for each query image, which is more efficient compared with building one single graph with all proposals and classes.

\section{Pairwise cosine similarity of all classes in MSCOCO}
\label{Paiwise_cosine_section}

We show in Fig.\ \ref{Paiwise_cosine} the pairwise cosine similarity of all 80 classes (including 60 base classes and 20 novel classes) in MSCOCO. We can find that there are some similar classes between the novel classes and base classes, for example, car and truck, chair and bench. This motivates us to `borrow' the robust features from these base classes to enhance novel class prototypes.

\end{document}